\documentclass[10pt,twocolumn,letterpaper]{article}

\usepackage{cvpr}
\usepackage{times}
\usepackage{epsfig}
\usepackage{graphicx}
\usepackage{amsmath}
\usepackage{amssymb}
\usepackage{enumitem}
\usepackage{mathtools}  % lVert, rVert
\usepackage{mwe}
\usepackage{placeins}  % floatbarrier
\usepackage{tabularx}
\usepackage{xcolor}
\usepackage[pagebackref=true,breaklinks=true,letterpaper=true,colorlinks,bookmarks=false]{hyperref}

\newcommand{\R}{\mathbb{R}}

\DeclareMathOperator*{\argmin}{arg\,min}
\DeclarePairedDelimiter\norm{\lVert}{\rVert_2}
\DeclarePairedDelimiter\abs{|}{|}
\graphicspath{ {images/} }

\cvprfinalcopy % *** Uncomment this line for the final submission

\def\cvprPaperID{5966} % *** Enter the CVPR Paper ID here

\ifcvprfinal\fi
\begin{document}

%%%%%%%%% TITLE
\title{RPM-Net: Robust Point Matching using Learned Features}
\author{Zi Jian Yew \qquad Gim Hee Lee \\
Department of Computer Science, National University of Singapore\\
{\tt\small \{zijian.yew, gimhee.lee\}@comp.nus.edu.sg}
% \and
}

\maketitle
%\thispagestyle{empty}

%%%%%%%%% ABSTRACT
\begin{abstract}
Iterative Closest Point (ICP) solves the rigid point cloud registration problem iteratively in two steps: (1) make hard assignments of spatially closest point correspondences, and then (2) find the least-squares rigid transformation. The hard assignments of closest point correspondences based on spatial distances are sensitive to the initial rigid transformation and noisy/outlier points, which often cause ICP to converge to wrong local minima. In this paper, we propose the RPM-Net -- a less sensitive to initialization and more robust deep learning-based approach for rigid point cloud registration. To this end, our network uses the differentiable Sinkhorn layer and annealing to get soft assignments of point correspondences from hybrid features learned from both spatial coordinates and local geometry. To further improve registration performance, we introduce a secondary network to predict optimal annealing parameters. Unlike some existing methods, our RPM-Net handles missing correspondences and point clouds with partial visibility. Experimental results show that our RPM-Net achieves state-of-the-art performance compared to existing non-deep learning and recent deep learning methods. Our source code is available at the project website\footnote{\url{https://github.com/yewzijian/RPMNet}}.
\end{abstract}

%%%%%%%%% BODY TEXT
\begin{figure}[ht]
\begin{center}
\includegraphics[width=\linewidth]{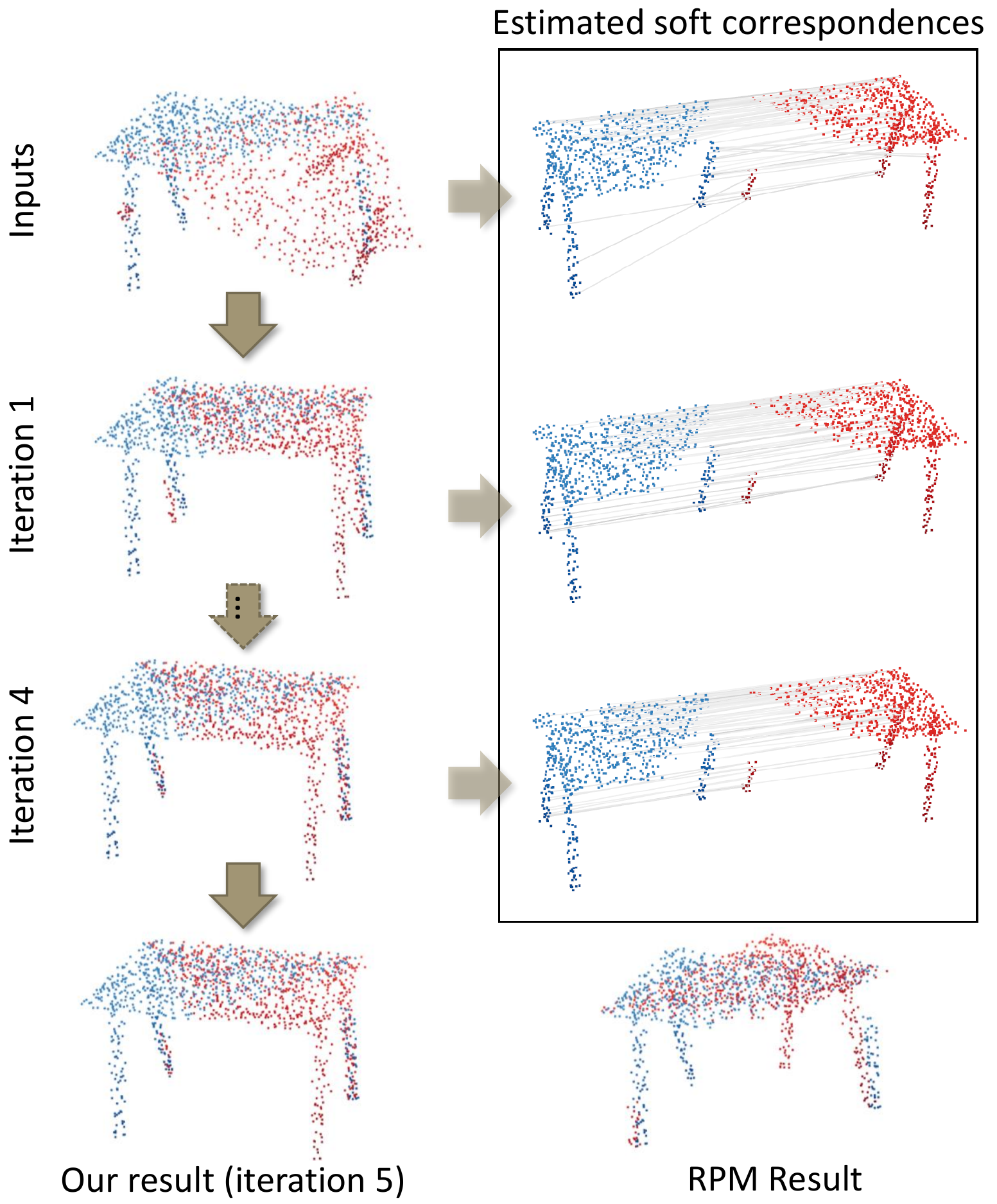}
\end{center}
\vspace{-2mm}
\caption{Our RPM-Net estimates soft correspondences from hybrid features learned from both spatial coordinates and local geometry of the points, and converges to the correct solution after 5 iterations. In contrast, RPM \cite{rpm} gets trapped in a local minima.}
\label{fig:teaser}
\end{figure}

\section{Introduction}
Rigid point cloud registration refers to the problem of finding the rigid transformation to align two given point clouds with unknown point correspondences. It has applications in many areas of computer vision and robotics, \eg robot and object pose estimation, point cloud-based odometry and mapping, etc.  Rigid point cloud registration is a chicken-and-egg problem that requires solving for both unknown point correspondences and rigid transformation to align the point clouds, and is thus commonly known as the simultaneous pose and correspondence problem \cite{3dreg-revisited}. Knowledge of either point correspondences or rigid transformation trivializes the problem.

ICP \cite{icp} is widely accepted to be the de facto algorithm for solving the rigid point cloud registration problem. It solves for both the point correspondences and rigid transformation by alternating between two steps: (1) assigns each point from the reference point cloud to its spatially closest point in the source point cloud, and (2) computes the least-squares rigid transformation between the correspondences. Unfortunately, ICP is highly sensitive to initialization and often converges to wrong local minima. It can also fail in the presence of noisy/outlier points. 
This limits ICP to relatively low noise and outlier-free scenarios with good initial rigid transforms, and precludes its use in applications such as registration of noisy scans and global registration. A recent deep learning-based ICP -- Deep Closest Point (DCP) \cite{deepcp} computes point correspondences from deep features to desensitize initialization, but remains not robust to outliers and does not work well on partially visible point clouds.

Many works \cite{chui-featuregmm,rpm,kernelcorr} are proposed to mitigate the problems of ICP, and
one prominent work is the Robust Point Matching (RPM) \cite{rpm}. It starts with soft assignments of the point correspondences, and gradually hardens the assignments through deterministic annealing. 
As we show in the experiments, although RPM is more robust than ICP, it remains sensitive to initialization and local minima as the point correspondences are still obtained solely from spatial distances.
On the other hand, feature-based methods \cite{FPFH,SHOT,USC} avoid initialization and the local minima problem by detecting distinctive keypoints and describing the local geometry of the keypoints using feature descriptors. Feature descriptors in one point cloud can then be matched to those in the other, and the rigid transformation can be solved robustly using a RANSAC scheme. However, these methods only work well for point clouds with distinctive geometric structures \cite{pcrnet}.

In this paper, we propose a deep learning-based RPM, the RPM-Net: an end-to-end differentiable deep network that preserves robustness of RPM against noisy/outlier points while desensitizing initialization with point correspondences from learned feature distances instead of spatial distances. To this end, we design a feature extraction network to compute hybrid features of each point from its spatial coordinates \emph{and} geometric properties, and then use a Sinkhorn \cite{sinkhorn} layer and annealing to get soft assignments of the point correspondences from these hybrid features. The fusion of spatial coordinates and geometric properties, and learning from data improve point correspondences. This desensitizes initialization and enhance the ability to register point clouds with missing correspondences and partial visibility. Similar to ICP and most of its variants, our RPM-Net refines rigid point cloud registration iteratively. Furthermore, we introduce a secondary network to predict optimal annealing parameters based on the current state of the alignments, \ie 
% Our annealing is non-deterministic since
our annealing does not follow a fixed schedule. Together with the use of hybrid features, our algorithm can converge in a small number of iterations as illustrated in the example shown in Figure \ref{fig:teaser}. Experiments show that our RPM-Net achieves state-of-the-art performance compared to existing non-deep learning and recent deep learning methods. Our main contributions are: \vspace{-0.1cm}
\begin{itemize}
    \item Learn hybrid features with a feature extraction network, Sinkhorn layer and annealing to desensitize initialization and enhance robustness of rigid point cloud registration.\vspace{-0.1cm}
    \item Introduce a secondary network to predict optimal annealing parameters. \vspace{-0.1cm}
    \item Suggest a modified Chamfer distance metric to improve measurement of registration quality in the presence of symmetry or partial visibility. \vspace{-0.1cm}
    \item Show state-of-the-art performance compared to other existing works on experimental evaluations under clean, noisy, and partially visible datasets. 
\end{itemize}

%-------------------------------------------------------------------------
\section{Related Work}

\paragraph{Feature-Based Methods.}
Feature-based methods tackle the registration problem in a two-step approach: (1) establish point correspondences between the two point clouds, and (2) compute the optimal transformation from these correspondences. The first step is non-trivial and requires well-designed descriptors to describe distinctive keypoints in order to match them between point clouds. A large variety of handcrafted 3D feature descriptors have been proposed and a comprehensive survey can be found in \cite{featdesc-review}. Generally, these descriptors accumulate measurements (typically number of points) into histograms according to their spatial coordinates \cite{3DSC,SpinImage,USC}, or their geometric attributes such as curvature \cite{LSP} or normals \cite{SHOT}. To orientate the spatial bins, most of these methods require a local reference frame (LRF) which is hard to obtain unambiguously, so other works \eg PFH \cite{PFH} and FPFH \cite{FPFH} design rotation invariant descriptors to avoid the need for a LRF. More recent works apply deep learning to learn such descriptors. One of the earliest such works, 3DMatch \cite{zeng20163dmatch}, voxelizes the region around each keypoint and compute descriptors with a 3DCNN trained using a contrastive loss. Voxelization results in a loss of quality, so later works such as PPFNet \cite{ppfnet} uses a PointNet \cite{pointnet,pointnet++} architecture to learn features directly from raw point clouds. In addition to predicting feature descriptors, 3DFeat-Net \cite{3dfeatnet} and USIP \cite{li2019usip} also learn to detect salient keypoints. The main problem with feature based methods is that they require the point clouds to have distinctive geometric structures. Additionally, the resulting noisy correspondences require a subsequent robust registration step (\eg RANSAC) which does not fit well into typical learning frameworks.

\vspace{-0.2cm}
\paragraph{Handcrafted Registration Methods.}
The original ICP algorithms \cite{icp,chen-medioni} circumvent the need for feature point matching by alternating between estimating point correspondences and finding the rigid transform that minimizes the point-to-point \cite{icp} or point-to-plane \cite{chen-medioni} error.
Subsequent works try to improve upon the convergence of ICP by \eg, selecting suitable points \cite{gelfand-stable-sampling,rusinkiewicz-normal-sampling} or weighting point correspondences \cite{godin-weights}. An overview of ICP variants can be found in \cite{rusinkiewicz-normal-sampling}. Nevertheless, most ICP variants still require relatively good initialization to avoid converging to bad local minima.
A notable exception, Go-ICP \cite{goicp} uses a branch-and-bound scheme to search for the globally optimal registration at the trade-off of much longer computation times. Alternatively, the basin of convergence of ICP can be widened using soft assignment strategies \cite{chui-featuregmm,rpm,kernelcorr}.  In particular, RPM \cite{rpm} uses a soft assignment scheme with a deterministic annealing schedule to gradually ``harden'' the assignment at each iteration. 
IGSP \cite{igsp} uses a different approach and measures the point similarity on a hybrid metric space with the spatial coordinates of the point and handcrafted BSC \cite{bsc} features. However, the authors do not learn the features, and have to handcraft the weighting scheme between the spatial and feature distances. Our work builds upon the iterative framework of RPM. However, we consider distances between learned hybrid features during its soft assignment stage. Moreover, we do not use a predefined annealing schedule, instead we let the network decide the best settings to use at each iteration.

\vspace{-0.3cm}
\paragraph{Learned Registration Methods.}
Recent works improve existing methods with deep learning. PointNetLK \cite{pointnetlk} utilizes PointNet \cite{pointnet} to compute a global representation of each point cloud, then optimizes the transforms to minimize the distances between the global descriptors in an iterative fashion analogous to the Lucas-Kanade algorithm \cite{lk, lk-20years}. Later, PCRNet \cite{pcrnet} improves the robustness against noise by replacing the Lucas-Kanade step with a deep network. Deep Closest Point \cite{deepcp} proposes a different approach. It extracts features for each point to compute a soft matching between the point clouds, before using a differentiable SVD module to extract the rigid transformation. They also utilize a transformer network \cite{transformer} to incorporate global and inter point cloud information when computing the feature representations. Although shown to be more robust than traditional methods, the above works cannot handle partial-to-partial point cloud registration. A concurrent work, PRNet \cite{prnet} incorporates keypoint detection to handle partial visibility.
Our work uses a simpler approach and is more similar to Deep Closest Point, but unlike \cite{deepcp}, our network is able to handle outliers and partial visibility through the use of Sinkhorn normalization \cite{sinkhorn} from RPM, and uses an iterative inference pipeline to achieve high precision.

%-------------------------------------------------------------------------
\section{Problem Formulation}
Given two point clouds: $\mathbf{X}=\{\mathbf{x}_j \in \R^3 \mid j = 1, ..., J\}$ and $\mathbf{Y}=\{\mathbf{y}_k \in \R^3 \mid k = 1, ..., K\}$, which we denote as the \emph{source} and \emph{reference}, respectively, 
our objective is to recover the unknown rigid transformation $\{\mathbf{R}, \mathbf{t}\}$. $\mathbf{R} \in SO(3)$ is a rotation matrix and $\mathbf{t} \in \R^3$ is a translation vector that align the two point clouds.
We assume the point normals can be easily computed from the points. Unlike the most recent deep learning-based related work \cite{deepcp}, we do not assume a one-to-one correspondence between points. The two point clouds can have different number of points, \ie, $J \neq K$ or cover different extents.

%-------------------------------------------------------------------------
\section{Background: Robust Point Matching}
As mentioned earlier, our work builds upon the framework of RPM \cite{rpm}. We briefly describe the algorithm in this section for completeness and interested readers are referred to \cite{rpm} for further details. We define a match matrix $\mathbf{M}=\{0,1\}^{J \times K}$ to represent the assignment of point correspondences, where each element
\begin{equation}
    m_{jk} = 
        \begin{cases}
        1 & \text{if point $\mathbf{x}_j$ corresponds to $\mathbf{y}_k$} \\
        0 & \text{otherwise}
        \end{cases}.
\end{equation}
Let us first consider the case where there is a one-to-one correspondence between the points. In this case, $\mathbf{M}$ is a square matrix. The registration problem can be formulated as finding the rigid transformation $\{\mathbf{R}, \mathbf{t}\}$ and correspondence matrix $\mathbf{M}$ which best maps points in $\mathbf{X}$ onto $\mathbf{Y}$, \ie,
\begin{equation}\label{eq:reg-cost}
    \argmin_{\mathbf{M}, \mathbf{R}, \mathbf{t}} \sum_{j=1}^{J}{\sum_{k=1}^{K}{
        m_{jk} (\norm{\mathbf{R}\mathbf{x}_j + \mathbf{t} - \mathbf{y}_k}^2 - \alpha)
    }}, \\
\end{equation}
subject to $\sum_{k=1}^{K} m_{jk} = 1, \forall j$,  $\sum_{j=1}^{J} m_{jk} = 1, \forall k$, and $m_{jk} \in \{0, 1\}, \forall jk$. The three constraints enforces $\mathbf{M}$ to be a permutation matrix.
$\alpha$ is a parameter to control the number of correspondences rejected as outliers: any pair of points $(\mathbf{x}_j, \mathbf{y}_k)$ with a distance $\norm{\mathbf{R}\mathbf{x}_j + \mathbf{t} - \mathbf{y}_k}^2 < \alpha$ is taken to be an inlier since setting $m_{jk}=1$ decreases the cost in Eq.~\ref{eq:reg-cost}.

In RPM, the permutation matrix constraint is relaxed to a doubly stochastic constraint, \ie, each $m_{jk} \in [0, 1]$. The minimization of Eq.~\ref{eq:reg-cost} is then solved using deterministic annealing that iterates between two steps: (1) softassign, and (2) estimation of the rigid transformation. The match matrix $\mathbf{M}$ is estimated in the softassign step. To this end, each element $m_{jk} \in \mathbf{M}$ is first initialized as follows:
\begin{equation}\label{eq:traditional-M}
    m_{jk} \leftarrow e^{-\beta (\norm{\mathbf{R}\mathbf{x}_j + \mathbf{t} - \mathbf{y}_k}^2 - \alpha)},
\end{equation}
where $\beta$ is the annealing parameter to be increased over each iteration step: small initial values of $\beta$ result in soft assignments which help avoid falling into local minima. As $\beta$ increases, $m_{jk} \rightarrow \{0,1\}$ and the match matrix $\mathbf{M}$ becomes closer to a permutation matrix.
Alternate row and column normalizations are then performed to satisfy the doubly stochastic constraints. This is due to a result from Sinkhorn \cite{sinkhorn}, which states that a doubly stochastic matrix can be obtained from any square matrix with all positive entries by repeated applications of alternating row and column normalizations. Note that the assignments are deterministic (hence the term \emph{deterministic} annealing). $\beta$ controls the ``hardness'' of correspondences in a deterministic fashion. This contrasts with simulated annealing methods \cite{simulated-annealing} where the decision of whether to accept a certain solution is a stochastic function of the temperature.

Once the correspondences are estimated, the rigid transformation $\{\mathbf{R}, \mathbf{t}\}$ can be computed. Various methods can be used for this purpose, we follow \cite{deepicp,deepcp} to compute $\{\mathbf{R}, \mathbf{t}\}$ using SVD (Section \ref{svd}) in this paper.
Lastly, when $J \neq K$ or in the presence of outlier non-matching points, the equality constraints on $\mathbf{M}$ in Eq.~\ref{eq:reg-cost} become inequality constraints, but can be converted back into an equality constraint by introducing slack variables:
\begin{equation}
    \sum_{k=1}^{K} m_{jk} \leq 1, \enspace\forall j \quad \rightarrow \quad \sum_{k=1}^{K+1} m_{jk}  = 1, \enspace\forall j,
\end{equation}
and likewise for the column constraints.
In practice, this is implemented by adding an additional row and column of ones to the input of Sinkhorn normalization, \ie, $\mathbf{M}_{J+1,K}, \mathbf{M}_{J,K+1}$.

\begin{figure*}[ht]
\begin{center}
\includegraphics[width=\textwidth]{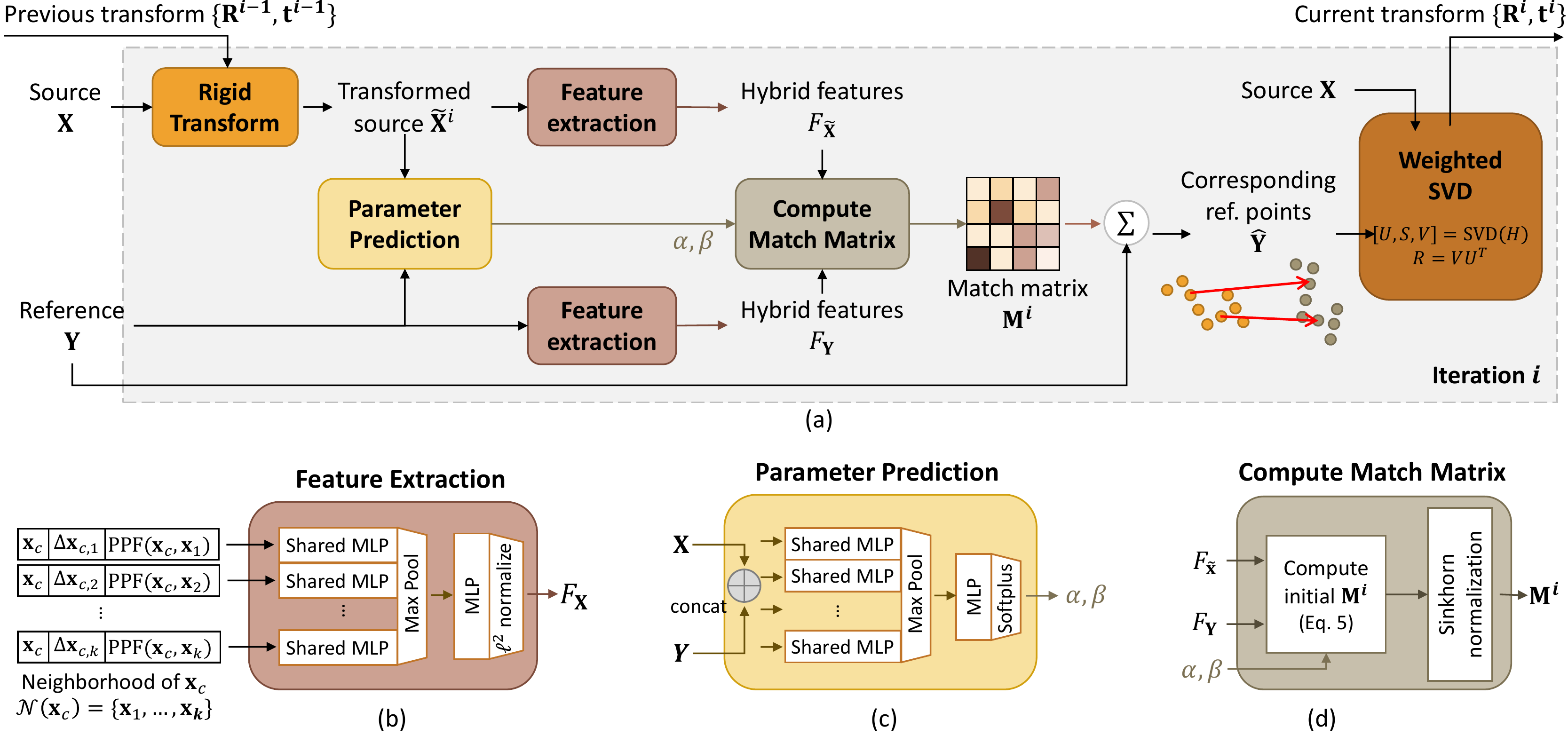}
\end{center}
\vspace{-2mm}
\caption{(a) Overview of our RPM-Net, (b) feature extraction network, (c) parameters prediction network, and (d) computation of match matrix $\mathbf{M}$. Superscripts denote the iteration count.}
\label{fig:network-architecture}
\end{figure*}

\section{Our RPM-Net}
Figure \ref{fig:network-architecture} shows an illustration of our RPM-Net..
We make two main changes to RPM: (1) spatial distances are replaced with learned hybrid feature distances, and (2) our network decides the values of $\alpha, \beta$ (\cf Eq. \ref{eq:traditional-M}) at each iteration.
At each iteration $i$, the source point cloud $\mathbf{X}$ is first transformed by the rigid transformation $\{{\mathbf{R}^{i-1}, \mathbf{t}^{i-1}}\}$ estimated from the previous step into the transformed point cloud $\tilde{\mathbf{X}}^i$. The feature extraction module (Section \ref{sect:feat-extraction}) then extracts hybrid features for the two point clouds.
Concurrently, a secondary parameter prediction network (Section \ref{sect:secondary-net}) predicts the optimal annealing parameters $\alpha, \beta$.
The hybrid features and $\alpha, \beta$ parameters are used to compute the initial match matrix, followed by Sinkhorn normalization to enforce the doubly stochastic constraints to get the final match matrix $\mathbf{M}^i$. Finally, the updated transformation $\{\mathbf{R}^i, \mathbf{t}^i\}$ is computed and 
used in the next iteration.

\subsection{Feature Extraction}\label{sect:feat-extraction}
We replace the spatial distances in Eq. \ref{eq:traditional-M} with distances between learned features, \ie, 
\begin{equation}\label{eq:learned-M}
    m_{jk} \leftarrow e^{-\beta (\norm{F_{\tilde{\mathbf{x}}_j} - F_{\mathbf{y}_k}}^2 - \alpha)},
\end{equation}
where $F_{\tilde{x}_j}$ and $F_{\mathbf{y}_k}$ are the features for points $\tilde{\mathbf{x}}_j \in \tilde{\mathbf{X}}^i$ and $\mathbf{y}_k \in \mathbf{Y}$, respectively.
Replacing spatial coordinates with learned features allows our algorithm to consider additional sources of information, \eg local geometric characteristics, during the computation of the assignments to avoid getting stuck in wrong local minima.

In our work, $F_{(\cdot)}$ is a hybrid feature containing information on both the point's spatial coordinates and local geometry. For a point $\mathbf{x}_c$ in either point cloud, we first define a local neighborhood $\mathcal{N}(\mathbf{x}_c)$ containing points within a distance of $\tau_{rad}$ from it. Its feature $F_{\mathbf{x}_c}$ is then given by:
\begin{equation}
    F_{\mathbf{x}_c} = f_\theta(\mathbf{x}_c, \{\Delta \mathbf{x}_{c, i}\}, \{\text{PPF}(\mathbf{x}_c, \mathbf{x}_i)\} ),
\end{equation}
where $f_\theta$ is a deep network parameterized by $\theta$, and $x_i \in \mathcal{N}(\mathbf{x}_c)$.
$\Delta \mathbf{x}_{c, i}$ denotes the neighboring points translated into a local frame by subtracting away the coordinates of the centroid point \cite{pointnet++}:
\begin{equation}
    \Delta \mathbf{x}_{c,i} = \mathbf{x}_i - \mathbf{x}_c.
\end{equation}
$\text{PPF}(\mathbf{x}_c, \mathbf{x}_i)$ are 4D point pair features (PPF) \cite{FPFH,ppfnet} that describe the surface between the centroid point $\mathbf{x}_c$ and each neighboring point $\mathbf{x}_i$ in a rotation invariant manner:
\begin{equation}
\begin{split}
    \text{PPF}(\mathbf{x}_c, \mathbf{x}_i) = (&\angle(\mathbf{n}_c, \Delta \mathbf{x}_{c,i}), 
                                               \angle(\mathbf{n}_i, \Delta \mathbf{x}_{c,i}), \\
                                              &\angle(\mathbf{n}_c, \mathbf{n_i}), 
                                               \norm{\Delta \mathbf{x}_{c,i}}),
\end{split}
\end{equation}
where $\mathbf{n}_c$ and $\mathbf{n}_i$ are the normals of points $\mathbf{x}_c$ and $\mathbf{x}_i$.
The above two inputs describe the local geometry, but do not contain information about the absolute positions of the points. In this work, we also include the absolute position of the centroid point $\mathbf{x}_c$. This gives our network the ability to refine the registration iteratively as in the original RPM.

We implement $f_\theta$ using a PointNet \cite{pointnet} which is able to pool an arbitrary number of orderless points into a single descriptor. Specifically, to obtain the feature for $\mathbf{x}_c$, we first concatenate the raw features of each neighboring point $\mathbf{x}_i \in \mathcal{N}(\mathbf{x}_c)$ into a 10-D input vector $[\mathbf{x}_c, \Delta \mathbf{x}_{c, i}, \text{PPF}(\mathbf{x}_c, \mathbf{x}_i)]$. We then feed them into a series of shared dense layers, a max-pooling and additional dense layers, followed by $\ell^2$ normalization to obtain a single feature vector $F_{\mathbf{x}_c}$.

\subsection{Parameter Prediction Network}\label{sect:secondary-net}
In the original RPM algorithm, the value of the outlier parameter $\alpha$ and the annealing schedule for $\beta$ (Eq. \ref{eq:learned-M}) are manually set for each dataset. These parameters are dataset dependent and have to be tuned for each dataset.
In our RPM-Net, these parameters are difficult to set manually since they are dependent on the learned features.
We argue that a fixed annealing schedule is unnecessary as the parameters can be chosen based on the current state of alignment instead of the iteration count. Correspondingly, we use a secondary network that takes both point clouds as input and predicts the parameters for the current iteration. In particular, we concatenate the two point clouds to form a $(J+K, 3)$ matrix, augment it with a fourth column containing 0 or 1 depending on which point cloud the point originates from, and feed it into a PointNet that outputs $\alpha$ and $\beta$. To ensure that the predicted $\alpha$ and $\beta$ are positive, we use a softplus activation for the final layer.

\subsection{Estimating the Rigid Transformation}\label{svd}
Once the soft assignments are estimated, the final step is to estimate the rigid transformation. For each point $\mathbf{x}_j$ in $\mathbf{X}$, we compute the corresponding coordinate in $\mathbf{Y}$:
\begin{equation}
    \hat{\mathbf{y}}_j = \frac{1}{\sum_{k}^{K}{m_{jk}}} \sum_{k}^{K}{m_{jk} \cdot \mathbf{y}_k}.
\end{equation}
We then follow \cite{deepicp,deepcp} and use the SVD to solve for the rigid transformation, which has been shown to be differentiable in \cite{svd-diff}. Since not every point $\mathbf{x}_j$ might have a correspondence, we weigh each correspondence $(\mathbf{x}_j, \hat{\mathbf{y}}_j)$ by $w_j=\sum_{k}^{K}{m_{jk}}$ when computing the rigid transformation.

\subsection{Loss Functions}
Our primary loss function is the $\ell^1$ distance between the source point cloud $\mathbf{X}$ transformed using the groundtruth transformation $\{\mathbf{R}_{gt} , \mathbf{t}_{gt}\}$ and the predicted transformation $\{\mathbf{R}_{pred}, \mathbf{t}_{pred}\}$ \cite{deepicp}:
\begin{equation}
    \mathcal{L}_{reg}=\frac{1}{J} \sum_j^J{\abs{
       (\mathbf{R}_{gt}\mathbf{x}_j + \mathbf{t}_{gt}) - 
       (\mathbf{R}_{pred}\mathbf{x}_j + \mathbf{t}_{pred})
    }}
\end{equation}
We notice empirically that the network has the tendency to label most points as outliers
with only the above registration loss. To alleviate this issue, we add a secondary loss on the computed match matrix $\mathbf{M}$ to encourage inliers:
\begin{equation}
    % \mathcal{L}_{inlier} = -\frac{1}{K}\sum_j^J{m_{jk}} - \frac{1}{J}\sum_k^K{m_{jk}}.
    \mathcal{L}_{inlier} = -\frac{1}{J} \sum_j^J \sum_k^K {m_{jk}} - \frac{1}{K} \sum_k^K \sum_j^J {m_{jk}}.
    % \mathcal{L}_{inlier} = -\frac{1}{\min(J, K)} \sum_j^J \sum_k^K {m_{jk}}.
\end{equation}
The overall loss is the weighted sum of the two losses:
\begin{equation}
    \mathcal{L}_{total} = \mathcal{L}_{reg} + \lambda \mathcal{L}_{inlier},
\end{equation}
where we use $\lambda=0.01$ in all our experiments. We compute the loss for every iteration $i$, but weigh the losses by $\frac{1}{2}^{(N_i-i)}$ to give later iterations higher weights, where $N_i$ is the total number of iterations during training.

\subsection{Implementation Details}
The overall network is implemented as a recurrent neural network with an inner loop for the Sinkhorn normalization.
We follow \cite{deeppermnet} in our implementation of the Sinkhorn normalization by unrolling it for a fixed number of steps (set to 5). 
Although gradients can flow from one iteration to the other, in practice, that does not improve performance and causes training instability. We adopt a simple solution of stopping the $\{\mathbf{R}, \mathbf{t}\}$ gradients at the start of each iteration. This means every iteration becomes independent and we can just execute one iteration during training. 
Nevertheless, we run $N_i=2$ iterations of alignment during training since this allows the network to see data with smaller misalignments more often and consequently learn how to refine the registration in subsequent iterations. During test time, we use $N_i=5$ iterations to achieve more precise registration.
For both feature extraction and parameter prediction networks, we use ReLU activation with group normalization \cite{groupnorm} on all layers except the last. Our feature extraction network considers a neighborhood of $\tau_{rad}=0.3$, and outputs features of dimension 96.
We train the network using ADAM optimizer \cite{adam} with a learning rate of 0.0001.

%-------------------------------------------------------------------------
\section{Experiments}
\subsection{ModelNet40 Dataset}
We evaluate our algorithm on the ModelNet40 \cite{modelnet} dataset, which contains CAD models from 40 man-made object categories. We make use of the processed data from \cite{pointnet}, which contains 2,048 points sampled randomly from the mesh faces and normalized into a unit sphere. The dataset contains official train/test splits for each category. To evaluate the ability of our network to generalize to different object categories, we use the train and test splits for the first 20 categories for training and validation respectively, and the test split of the remaining categories for testing. This results in 5,112 train, 1,202 validation, and 1,266 test models.
Following \cite{deepcp}, we sample rotations by sampling three Euler angle rotations in the range $[0, 45^{\circ}]$ and translations in the range $[-0.5, 0.5]$ on each axis during training and testing. We transform the source point cloud $\mathbf{X}$ using the sampled rigid transform and the task is to register it to the unperturbed reference point cloud $\mathbf{Y}$.

\subsection{Evaluation Metrics}
We evaluate the registration by computing the mean isotropic rotation and translation errors:
\begin{align}
    \text{Error}(\mathbf{R}) = \angle (\mathbf{R}_{GT}^{-1} \hat{\mathbf{R}} ),\quad
    \text{Error}(\mathbf{t}) = \norm{\mathbf{t}_{GT} - \hat{\mathbf{t}}},
\end{align}
where $\{\mathbf{R}_{GT}, \mathbf{t}_{GT}\}$ and $\{\hat{\mathbf{R}}, \hat{\mathbf{t}}\}$ denote the groundtruth and estimated transformation, respectively. $\angle(\mathbf{X}) = \arccos(\frac{tr(\mathbf{X})-1}{2})$ returns the angle of rotation matrix $\mathbf{X}$ in degrees.
For consistency with previous work \cite{deepcp}, we also provide the mean absolute errors over euler angles and translation vectors. Note however that these metrics are anisotropic.

The above metrics unfairly penalizes the alignment to an alternative solution in the case of symmetry commonly found in ModelNet40 models, so we also propose a modified Chamfer distance metric between the transformed source point cloud $\mathbf{X}$ and the reference point cloud $\mathbf{Y}$:
\begin{equation}
\begin{split}
    \tilde{CD}(\mathbf{X}, \mathbf{Y}) &= 
       \frac{1}{\abs{\mathbf{X}}} 
       \sum_{\mathbf{x} \in \mathbf{X}}{\min_{\mathbf{y} \in \mathbf{Y}_{\text{clean}}}{\norm{\mathbf{x-y}}^2}} + \\ & \qquad \qquad
       \frac{1}{\abs{\mathbf{Y}}} 
       \sum_{\mathbf{y} \in \mathbf{Y}}{\min_{\mathbf{x} \in \mathbf{X}_{\text{clean}}}{\norm{\mathbf{x-y}}^2}},
\end{split}
\end{equation}
where we modified the Chamfer distance to compare with the \emph{clean} and \emph{complete} versions of the other point cloud.

\subsection{Baseline Algorithms}
We compare the performance of our RPM-Net with the following handcrafted registration algorithms: ICP \cite{icp}, FGR \cite{fgr}, and RPM \cite{rpm}, as well as recent deep learning based registration works: PointNetLK \cite{pointnetlk} and Deep Closest Point (DCP-v2) \cite{deepcp}.
We use the implementations of ICP and FGR in Intel Open3D \cite{open3d}, and our own implementation of RPM. For PointNetLK and Deep Closest Point, we use the implementation provided by the authors but retrained the networks since both works do not provide the required pretrained models\footnote{Deep Closest Point provides pretrained models but not for matching of unseen categories and noisy data.}.

\subsection{Clean Data}
We follow the protocol in \cite{deepcp} and evaluate the registration performance on the clean data. We randomly sample the same 1,024 points for the source and reference point clouds from the 2,048 points in ModelNet40 dataset, and then apply a random rigid transformation on the source point cloud and shuffle the point order. Under this setting, each point in the source point cloud $\mathbf{X}$ has a exact correspondence in the reference point cloud $\mathbf{Y}$. All learned models including ours are trained on the clean data.
Table \ref{table:clean-performance} shows the performance of the various algorithms on clean data. Our method achieves very accurate registration and ranks first or second in all measures. It outperforms all learned and handcrafted methods except FGR. However, as we will see in subsequent sections, FGR is highly susceptible to noise.
A qualitative comparison of the registration results can be found in Figure \ref{fig:qualitative}(a).

\begin{table}
\small
\begin{center}
\setlength\tabcolsep{4.5pt}
\begin{tabularx}{\linewidth}{X | c c c c c}
  \hline
  Method & \multicolumn{2}{c}{Anisotropic err.} & \multicolumn{2}{c}{Isotropic err.} & $\tilde{CD}$\\
   & (Rot.) & (Trans.) & (Rot.) & (Trans.) &  \\
  \hline
  ICP         &  3.114  &  0.02328  &  6.407  &  0.0506  &  0.002975  \\
  RPM         &  1.121  &  0.00636  &  2.426  &  0.0141  &  0.000360  \\
  FGR         &  \textbf{0.010}  &  \textbf{0.00011}  &  \textbf{0.022}  &  \textbf{0.0002}  &  \emph{0.000012}  \\
  PointNetLK  &  0.418  &  0.00241  &  0.847  &  0.0054  &  0.000138  \\
  DCP-v2      &  2.074  &  0.01431  &  3.992  &  0.0292  &  0.001777  \\
  \hline
  Ours        &  \emph{0.028}  &  \emph{0.00016}  &  \emph{0.056}  &  \emph{0.0003}  &  \textbf{0.000003} \\
  \hline
\end{tabularx}
\end{center}
\caption{Performance on Clean Data. Bold and italics denote best and second best performing measures. Note: DCP-v2's results are based on our trained model and are marginally worse than its reported \cite{deepcp} performance of an anisotropic error of $2.007^\circ$ (rot) and 0.0037 (trans).}
\label{table:clean-performance}
\end{table}

\subsection{Gaussian Noise} \label{sect:Gaussian-noise}
In this experiment, we evaluate the performance in the presence of noise and sampling differences, which are present in real world point clouds. We randomly sample 1,024 points from the models, but \emph{independently} for the source and reference point clouds. After applying the random rigid transform to the source point cloud, we randomly jitter the points in both point clouds by noises sampled from $\mathcal{N}(0, 0.01)$ and clipped to [-0.05, 0.05] on each axis. This experiment is significantly more challenging due to noise and non one-to-one correspondences. 
We train all learned models on the noisy data with the exception of PointNetLK, which we reuse the model from the previous section since that gives better performance. The results are shown in Table \ref{table:noisy-performance}. 
Our network outperforms all handcrafted and learned methods as we explicitly handle for points with no correspondences. On the other hand, Deep Closest Point requires every point to have a correspondence and does not perform well when this condition is violated.
A qualitative example of registration on noisy data can be found in Figure \ref{fig:qualitative}(b).

\begin{table}
\small
\begin{center}
\setlength\tabcolsep{4.5pt}
\begin{tabularx}{\linewidth}{X | c c c c c}
  \hline
  Method & \multicolumn{2}{c}{Anisotropic err.} & \multicolumn{2}{c}{Isotropic err.} & $\tilde{CD}$\\
   & (Rot.) & (Trans.) & (Rot.) & (Trans.) &  \\
  \hline
  ICP         &  3.414  &  0.0242  &  6.999  &  0.0514  &  0.00308  \\
%   RPM         &  3.134  &  0.0226  &  6.257  &  0.0471  &  0.00254  \\
  RPM         &  1.441  &  0.0094  &  2.994  &  0.0202  &  0.00083  \\
  FGR         &  1.724  &  0.0120  &  2.991  &  0.0252  &  0.00130  \\
  PointNetLK  &  1.528  &  0.0128  &  2.926  &  0.0262  &  0.00128  \\
  DCP-v2      &  4.528  &  0.0345  &  8.922  &  0.0707  &  0.00420  \\
  \hline
  Ours        & \textbf{0.343}      & \textbf{0.0030}       & \textbf{0.664}      & \textbf{0.0062}       &   \textbf{0.00063} \\
  \hline
\end{tabularx}
\end{center}
\caption{Performance on data with Gaussian noise. The Chamfer distance using groundtruth transformations is 0.00055.}
\label{table:noisy-performance}
\end{table}

\subsection{Partial Visibility}
\begin{figure*}[ht]
\begin{center}
\includegraphics[width=\textwidth]{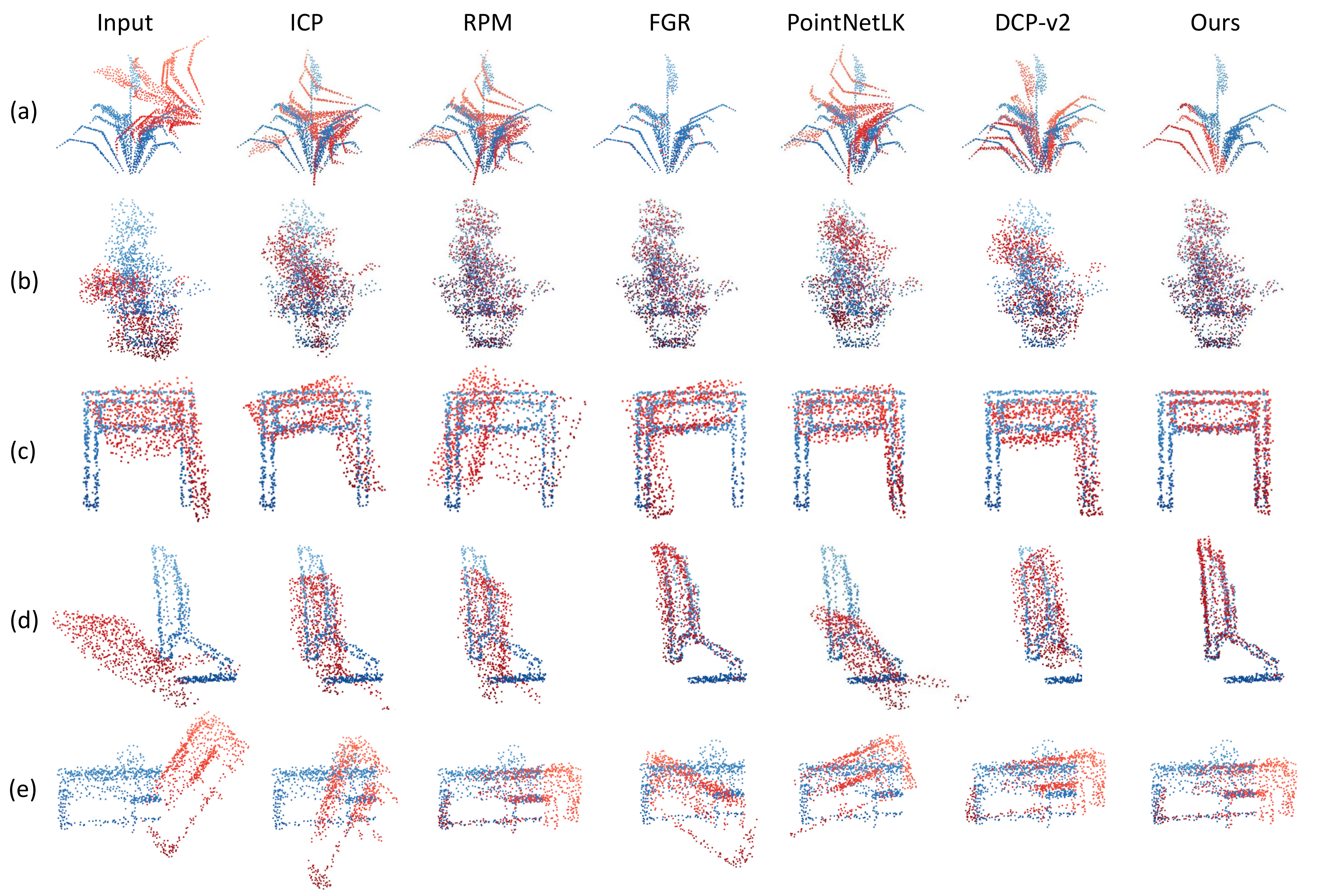}
\end{center}
\vspace{-5mm}
\caption{Qualitative registration examples on (a) Clean data, (b) Noisy data, and (c, d, e) Partially visible data.} 
\label{fig:qualitative}
\end{figure*}

We evaluate the performance on partially visible point clouds, where the two point clouds do not fully overlap in extent.
Depending on the acquisition method, real world point cloud data are often partial, \eg RGD-D scans contain only points that are visible to the camera. Consequently, handling partial point clouds is an important requirement for many applications. 
We simulate partial visibility in the following manner. For each point cloud, we sample a half-space with a random direction $\in \mathcal{S}^2$ and shift it such that approximately 70\% of the points are retained. Similar to the previous section, points are jittered and sampled independently. For this experiment, we downsample to 717 points instead of 1,024 to maintain a similar point density as the previous sections.
We train DCP-v2 and our method on the partially visible data. For PointNetLK, we adopt similar procedure suggested by the authors to sample visible points in the other point cloud. However, we sample visible points from the points with a respective nearest point in the other point cloud within a distance of $\tau=0.02$ during each iteration
since only the partial point cloud is available for both point clouds in our setting. This procedure improves inference performance but did not work well during training. Consequently, we continue to use the clean model of PointNetLK in this experiment. 
Table \ref{table:crop-performance} shows the performance on partially visible data. Our approach significantly outperforms all baseline methods. Interestingly, despite our best efforts at tuning the parameters in RPM, it also performs poorly. Since both RPM and our approach share a similar scheme for rejecting outliers, this highlights the benefit of our learned feature distances.
Example results on partially visible data are shown in Figures \ref{fig:teaser} and \ref{fig:qualitative}(c-e).

\begin{table}
\small
\begin{center}
\setlength\tabcolsep{4.5pt}
\begin{tabularx}{\linewidth}{X | c c c c c}
  \hline
  Method & \multicolumn{2}{c}{Anisotropic err.} & \multicolumn{2}{c}{Isotropic err.} & $\tilde{CD}$\\
   & (Rot.) & (Trans.) & (Rot.) & (Trans.) &  \\
  \hline
  ICP         &  13.719  &  0.132  &  27.250  &  0.280  &  0.0153  \\
  RPM         &   9.771  &  0.092  &  19.551  &  0.212  &  0.0081  \\
%   RPM         &  25.519  &  0.190  &  45.882  &  0.398  &  0.0204  \\
  FGR         &  19.266  &  0.090  &  30.839  &  0.192  &  0.0119  \\
  PointNetLK  &  15.931  &  0.142  &  29.725  &  0.297  &  0.0235  \\
  DCP-v2      &  6.380   &  0.083  &  12.607  &  0.169   &  0.0113  \\
  \hline
  Ours        & \textbf{0.893}      & \textbf{0.0087}       & \textbf{1.712}      & \textbf{0.018}       &   \textbf{0.00085} \\
\hline
\end{tabularx}
\end{center}
\caption{Performance on partially visible data with noise. The Chamfer distance using groundtruth transformations is 0.00055.}
\label{table:crop-performance}
\end{table}

\subsection{Ablation Studies}
We perform ablation studies to better understand how various choices affect the performance of the algorithm. All studies in this section are evaluated on the partial visibility setting, and we only show the isotropic and Chamfer distance metrics for conciseness.

\vspace{-3mm}
\paragraph{Effects of different components.}
Comparing rows 1-3 and 5 of Table \ref{table:ablation}, we observe that all of $\mathbf{x}$, $\Delta \mathbf{x}$ and PPF are required to achieve the highest performance.
Excluding the absolute positions of the points (row 3) results in a significant drop in performance. This indicates the importance of considering point positions at each iteration when performing iterative refinement.
It is also noteworthy that even without PPF features (row 1), the algorithm still outperforms DCP-v2. This is despite DCP-v2 using a more sophisticated Dynamic Graph CNN \cite{dgcnn} architecture and an attention \cite{transformer} network. We attribute this to our outlier handling and iterative registration scheme. 
To understand the importance of our parameter prediction network, we lastly compare with a variant of our network in Table \ref{table:ablation} (row 4) where we replace our parameter prediction network with two learnable parameters for $\alpha$ and $\beta$. These parameters are trained with the network weights, and the same values are used for each iteration. We can see that the learned annealing schedule from the parameter prediction network improves registration performance.

\begin{table}[t]
\small
\begin{center}
\setlength\tabcolsep{3pt}
\begin{tabularx}{0.81\linewidth}{c c c | c | c c c }
  \hline
  $\mathbf{x}$ & $\Delta\mathbf{x}$ & $PPF$&  Anneal    &  \multicolumn{2}{c}{Isotropic err.}  &  $\tilde{CD}$ \\
             &              &             &             &  (Rot.) & (Trans.) & \\
  \hline
  \checkmark & \checkmark   &             & \checkmark  &  3.302  &  0.0350  &  0.00153 \\
  \checkmark &              & \checkmark  & \checkmark  &  2.781  &  0.0273  &  0.00123 \\
             & \checkmark   & \checkmark  & \checkmark  &  5.501  &  0.0496  &  0.00351 \\
  \checkmark & \checkmark   & \checkmark  &             &  2.220  &  0.0238  &  0.00103 \\
  \hline
  \checkmark & \checkmark   & \checkmark  & \checkmark  &  \textbf{1.712}  &  \textbf{0.0183}  &  \textbf{0.00085} \\
  \hline
\end{tabularx}
\end{center}
\caption{Effects of each component on the registration performance. $\mathbf{x}$ and $\Delta \mathbf{x}$ denote 
the absolute centroid center coordinates and the local coordinates of the neighboring points respectively.}
\label{table:ablation}
\end{table}

\begin{figure}[t]
\begin{center}
\vspace{-1mm}
\includegraphics[width=0.88\linewidth]{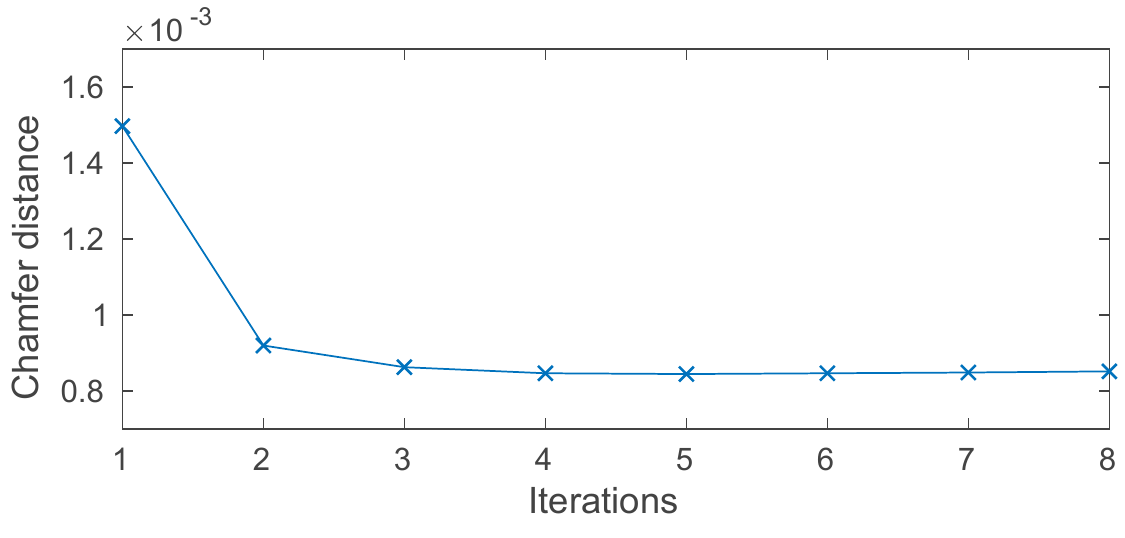}
\end{center}
\vspace{-4mm}
\caption{Chamfer distance over registration iterations. The initial average Chamfer distance of the inputs is 0.1899, which is not shown in graph for clarity.}
\label{fig:iterations}
\end{figure}

\vspace{-3mm}
\paragraph{How many iterations are needed?}
Figure \ref{fig:iterations} shows the average Chamfer distance after each iteration. Most performance gains are in the first two iterations, and the registration mostly converges after 5 iterations (which we use for all experiments).

\subsection{Computational Efficiency}\label{efficiency}
We compare the inference time of various algorithms in Table \ref{table:timings}, averaged over the entire test set. We perform this experiment on a 3.0GHz Intel i7-6950X and a Nvidia Titan RTX. For our algorithm, we provide the timings for 5 iterations as used in previous experiments. Note that ICP and FGR are executed on CPU and the remaining algorithms on a GPU.
Our algorithm is significantly faster than RPM which requires a large number of iterations. It is however slower than ICP as well as the non-iterative DCP-v2.

\begin{table}[ht]
\small
\begin{center}
\setlength\tabcolsep{4.5pt}
\begin{tabularx}{\linewidth}{X | c c c c c c}
  \hline
  \# points & ICP & RPM  & FGR & PointNetLK & DCP-v2 & Ours\\
  \hline
  512       &  8  & 66   & 22  &  161       &  5     & 25 \\
  1024      & 18  & 144  & 84  &  176       &  9     & 52 \\
  2048      & 28  & 447  & 148 &  209       &  21    & 178 \\
  \hline
\end{tabularx}
\end{center}
% Batch sizes: PointNetLK: 8, Ours: 8 or 16 (no diff)
\caption{Average time required for registering a point cloud pair of various sizes (in milliseconds).}
\vspace{-2mm}
\label{table:timings}
\end{table}

%-------------------------------------------------------------------------
\section{Conclusion}
We present the RPM-Net for rigid point cloud registration. Our approach is a deep learning-based RPM that desensitizes initialization and improves convergence behavior with learned fusion features. Furthermore, the use of the differentiable Sinkhorn normalization with slack variables to enforce partial doubly stochastic constraints allows our method to explicitly handle outliers. We also propose a secondary network to predict optimal annealing parameters that further improves performance.
Experimental results show our method yields state-of-the-art performance on the ModelNet40 dataset over various evaluation criteria.

\vspace{-2mm}
\paragraph{Acknowledgement.}
This work was partially supported by the Singapore MOE Tier 1 grant R-252-000-A65-114.

%-------------------------------------------------------------------------

\FloatBarrier

{\small
\bibliographystyle{ieee_fullname}
% \bibliography{references}

}

\end{document}